\documentclass[10pt]{article}
\newif\ifblindreview

\usepackage[letterpaper]{geometry}
\usepackage{amta2024}
\usepackage{times}
\usepackage{url}
\usepackage{latexsym}
\usepackage{natbib}
\usepackage{layout}
\usepackage{multicol}
\usepackage{booktabs,array}
\usepackage{float}
\usepackage[hidelinks]{hyperref}
\usepackage[all]{hypcap}
\usepackage{graphicx}
\usepackage{inconsolata}
\usepackage{xcolor}
\hypersetup{
    colorlinks,
    linkcolor={red!50!black},
    citecolor={blue!50!black},
    urlcolor={blue!80!black}
}
\usepackage{amsmath}
\usepackage{xcolor}%
\usepackage{textcomp}%
\usepackage{manyfoot}%
\usepackage{booktabs}
\usepackage[utf8]{inputenc}
\usepackage{algorithm}
\usepackage{algpseudocode}
\usepackage{listings}%
\usepackage{caption}
\usepackage{enumitem}

\usepackage{multirow}
\usepackage{boldline}
\usepackage{adjustbox}
\usepackage{colortbl}
\definecolor{lightgray}{gray}{0.9}
\usepackage{siunitx}
\usepackage{booktabs}
\ifblindreview
  \newcommand{\authorinfo}{\author{}} % Leave this blank!
\else
  \newcommand{\authorinfo}{% Add your author information here:
    \author{
      \name{\bf Javad Pourmostafa Roshan Sharami} \hfill \addr{j.pourmostafa@tilburguniversity.edu}\\
      \name{\bf Dimitar Shterionov} \hfill \addr{d.shterionov@tilburguniversity.edu}\\
      \name{\bf Pieter Spronck} \hfill \addr{p.spronck@tilburguniversity.edu}\\
      \addr{\small Department of Cognitive Science and Artificial Intelligence, Tilburg University%, the Netherlands
      }
    }
  }
\fi

\begin{document}

\amtaHeader{x}{x}{xxx-xxx}{2015}{45-character paper description goes here}{Author(s) initials and last name go here}
\title{\bf Guiding In-Context Learning of LLMs \\through Quality Estimation for Machine Translation}
\authorinfo

\maketitle
\pagestyle{empty}

\begin{abstract}
\vspace{8pt}
The quality of output from large language models (LLMs), particularly in machine translation (MT), is closely tied to the quality of in-context examples (ICEs) provided along with the query, i.e., the text to translate. The effectiveness of these ICEs is influenced by various factors, such as the domain of the source text, the order in which the ICEs are presented, the number of these examples, and the prompt templates used. Naturally, selecting the most impactful ICEs depends on understanding how these affect the resulting translation quality, which ultimately relies on translation references or human judgment. This paper presents a novel methodology for in-context learning (ICL) that relies on a search algorithm guided by domain-specific quality estimation (QE). Leveraging the XGLM model, our methodology estimates the resulting translation quality without the need for translation references, selecting effective ICEs for MT to maximize translation quality. % Our study also includes a comparison between fine-tuning a pre-trained language model (PLM) tailored to MT (mBART-50) and the proposed ICL approach. The study results demonstrate a significant performance enhancement compared to existing ICL methods in the literature. Additionally, our experiments show that while fine-tuning mBART-50 incurs significant computational costs and results in a notably lower BLEU score compared to our proposed ICL approach, it leads to improved contextual translation performance.
Our results demonstrate significant improvements over existing ICL methods and higher translation performance compared to fine-tuning a pre-trained language model (PLM), specifically mBART-50. %The script for running our experiments is publicly available at \url{anynomous.com}.

%Additionally, we investigate the impact of reordering ICEs based on unigram sequences when inputting them into XGLM. 
%Furthermore, our analysis reveals that although the translation quality can potentially be enhanced by ordering ICEs based on their unigram overlap %with the translated source, 
%this improvement lacks statistical significance. 

\end{abstract}

\begin{multicols}{2}

\section{Introduction}\label{sec:introduction}
Pre-trained large language models~(LLMs) quickly gained popularity (and continue to do so) due to their performance on a large set of natural language processing (NLP) tasks, including machine translation (MT)~\citep{zhu2023multilingual,xu2024paradigm}. However, the accuracy of their outputs is significantly influenced by the quality of the \emph{in-context} examples (ICEs) provided to them~\citep{10.1162/tacl_a_00324,alves-etal-2023-steering}.\footnote{For simplicity, we sometimes refer to it as ``example(s)'' throughout this paper.}
%The accuracy of outputs generated by pre-trained large language models~(LLMs) in Machine Translation~(MT) and other natural language processing (NLP) tasks~\cite{zhu2023multilingual,xu2024paradigm}, is significantly influenced by the quality of the examples provided to them, also known as in-context examples (ICEs)\footnote{For simplicity, we sometimes refer to it as ``example(s)'' throughout this paper.}~\cite{10.1162/tacl_a_00324}.
If these examples %are not appropriately selected –– that is, if they 
do not align well with the specific task and source domain, the LLMs' outputs can be inaccurate. Therefore, there is a critical need to develop (better) methods for selecting appropriate examples that match the task and source domain being translated. These methods collectively fall under the umbrella of in-context learning (ICL)~\citep{liu-etal-2022-makes}. %This is particularly critical, given the limited number of ICEs that can be inputted into the LLMs~\cite{agrawal-etal-2023-context}.

% LLMs have achieved unprecedented results on many (NLP) tasks, including machine translation (MT), a.o.~\cite{zhu2023multilingual,xu2024paradigm}. %And while prompting is an important factor on this quality~\cite{zhang2023prompting}, the used ICEs have also a substantial, if not even larger, impact.
Traditionally, creating ICEs for MT involves either random selection~\citep{sia-duh-2023-context} or using a strategy such as maximizing an evaluation metric like BLEU, to choose examples that improve the metric~\citep{agrawal-etal-2023-context}. %from the development set 
 %(i.e., task-level prompts)
 %This strategy is based on the rationale that development sets closely resemble the domain of the test dataset. However, such approaches may not be practical in real-world MT scenarios, as labeled development sets are not always accessible. 
The former was initially used for its simplicity and ease of implementation. However, relying on randomness %in these methods 
can lead to inconsistent results and pose significant computational costs~\citep{lu-etal-2022-fantastically}. %For example, a randomized technique, such as the task-level prompt, typically demands nearly 100 random trials and approximately 78 hours of inferring, which can be costly in terms of both computational resources and time. %to achieve translation results comparable to those in the existing literature. 
Recent state-of-the-art (SOTA) ICL approaches focus on retrieving training examples that are closely relevant to the context of source sentences of test sets using unsupervised retrievers, such as BM25~\citep{10.1561/1500000019}.
% This goes to related work
%accomplish this either through the embedding space of a pre-trained language model (PLM) –– a method previously shown to be effective in capturing context and domain~\cite{aharoni-goldberg-2020-unsupervised} –– or via unsupervised retrievers such as BM25 ~\cite{10.1561/1500000019} to provide additional context to the LLMs~\cite{shin-etal-2021-constrained,das-etal-2021-case,rubin-etal-2022-learning,agrawal-etal-2023-context}.
Recent %SOTA ICL 
studies have also shown that a range of factors, such as order~\citep{lu-etal-2022-fantastically}, template~\citep{10.1162/tacl_a_00324}, domain, and number of ICEs, significantly impact the performance~\citep{agrawal-etal-2023-context, raunak-etal-2023-dissecting}. %However, these factors may vary depending on the specific LLM being used.
%Naturally, these factors have different impacts on different LLMs, making it essential to systematically analyze each phenomenon for individual LLMs and source texts. %Importantly, their impact can vary even within individual source texts within a test set. %For example, %using XGLM~\cite{lin-etal-2022-shot}, 
%a source sentence with 16 ICEs may yield a BLEU score of 50, while the same sentence with only one ICE could achieve a BLEU score of 100. %
%Consequently, determining the optimal quantity and quality of ICEs, their order, and templates necessitates systematically analyzing each phenomenon for individual LLMs and translated sources. 

Naturally, the most effective ICEs for a given source text are the ones that would impact the resulting translation quality, which would ultimately depend on translation references or human judgment. %Notably, these ICE-related factors differ among LLMs, necessitating a systematic analysis tailored to individual LLMs and source texts. Nonetheless, carrying out such an analysis poses significant challenges, requiring substantial time and computational resources due to the rapid growth and diversity of LLMs in the field.
%Nevertheless, such an undertaking would be not only time-consuming but also computationally expensive, given the rapid growth and diversity of LLMs.
%However, selecting the right options, i.e., quality, quantity, etc., boils down to knowing how they would impact the resulting translation quality, which would ultimately depend on translation references or human judgment.
In MT, quality estimation~(QE) has become a standard approach for evaluating an MT system's output without relying on reference translations~\cite{blain-etal-2023-findings}.
Recently, \citet{lee-2020-two}, \citet{10.1007/978-981-99-7894-6_7}, and \citet{sharami-etal-2023-tailoring} showed the effectiveness of domain-specific QE when it comes to domain-specific MT (in contrast to the ineffectiveness of generic QE). Building on this and to address the aforementioned challenges, our work proposes to leverage domain-specific QE to assist in the selection of ICEs, with the goal of determining the suboptimal number and combination of ICEs to maximize MT quality, all without reference translations.
%We propose that domain-specific QE models, recently shown to be effective~\cite{lee-2020-two,sharami-etal-2023-tailoring}, could also be integrated into ICL methods to evaluate the quality of selected ICEs, including determining the optimal number of ICEs, without requiring translation references. 
As QE would assess the impact of different ICE combinations and sequences, we hypothesize that this integration has the potential to not only improve translation performance but also reduce processing time, as QE could result in smaller sets of ICEs, which would reduce the inference times~\citep{petrov2023language}. This is particularly crucial considering the limited number of ICEs that can be fed into LLMs~\citep{agrawal-etal-2023-context}. Therefore, our study aims to investigate the feasibility of selecting ICEs on a per-source basis. Specifically, we aim to answer the following research question (RQ): \textit{How effective are domain-specific QE models in determining ICEs for translation tasks in an LLM?}

%In addition, we follow two objectives: (i) Given the success of ordering ICEs based on their n-gram overlap match with the source, as per~\cite{agrawal-etal-2023-context}, we aim to investigate the efficacy of this approach within our proposed methodology. (ii) We aim to evaluate the impact of using our proposed methodology for ICL in comparison to fine-tuning a pre-trained multilingual MT model with regard to translation quality and computational costs. By considering all computational factors, this comparison allows us to determine whether ICL presents a more advantageous approach compared to fine-tuning a pre-trained MT model.  
    
Our proposed ICL methodology for MT combines an unsupervised retriever to select ICEs with QE to assess their impact on the translation quality, determining which ICE combination to include. Instead of feeding all selected examples, we only select examples whose QE points to maximizing the LLM translation quality. %In our proposed approach, we pass the examples with their respective order determined by the retriever module to the LLM. After obtaining the translated output from the LLM, along with the source, we feed them to the QE model and keep a record of scores estimated by the QE model. Finally, we select the combination of ICEs for each source, yielding the highest estimated BLEU score. 

Our findings on German-English translations demonstrate that our proposed approach outperforms the current SOTA ICL methods for MT as well as a fine-tuned mBART-50~\citep{tang2020multilingual}.

\section{ICL Using Quality Estimation for MT}\label{sec:methodology}
To utilize LLMs for effective MT, as noted in Section~\ref{sec:introduction}, what is needed is a set of examples to provide the context (and thus guide or steer the LLM toward a correct, context-specific translation) –– that is, a set of ICEs –– and what is further important is the number of ICEs and their combination.\footnote{The question of the order of examples is not specifically discussed in this paper but is left for future work.} Ultimately, what is required is that the ICEs provide context that is neither too specific nor too broad and can effectively boost the translation. Our goal with this work is to develop a methodology that optimizes both these aspects in order to deliver high-quality MT. Our methodology for identifying effective ICEs involves two key components: (1) an unsupervised retriever that locates examples closely related to the sentence to be translated and (2) a search algorithm that uses QE to select a combination of examples that leads to the improvement of translation quality, i.e., aiming to maximize the BLEU score. %The way these two are connected is as follows: the retrieved examples (initial ICEs) are fed into the search algorithm. Subsequently, the search algorithm determines the optimal combination of ICEs for each source text based on QE analysis, aiming to maximize the BLEU score. %we present details in Section~\ref{retrieval_module}

\subsection{Unsupervised Retriever Ranking}\label{retrieval_module}
We employ the \emph{BM25} ranking algorithm~\citep{10.1145/2682862.2682863} due to the effective utilization of unsupervised retriever methods demonstrated in previous research, such as \citep{agrawal-etal-2023-context}. BM25 sorts training pairs (source text and their translations) based on their relevance to a given query, i.e. the sentence to be translated.
%source text from the test set.
Subsequently, we select the top $K$ sentence pairs ranked by the algorithm, where $K$ is a hyperparameter that controls the number of pairs to be fed into the search algorithm. 
%A detailed explanation of this module's implementation is provided in Section~\ref{retrieval_module}.
%we pass the examples with their respective order determined by the retriever module to the LLM. After obtaining the translated output from the LLM, along with the source, we feed them to the QE model and keep a record of scores estimated by the QE model. Finally, we select the combination of ICEs for each source, yielding the highest estimated BLEU score. 
\subsection{Search Algorithm Coupled with QE}\label{sec:search_alg}
%We propose an algorithmic approach that combines a search algorithm with QE to identify the most effective combination of ICEs. Our method focuses on selecting ICEs that contribute to enhancing translation quality from a predefined set of ICEs provided by the BM25 algorithm. %We incorporate early stopping patience into the search process to manage computational costs.
Our search algorithm comprises three main phases: \emph{Selection, Translation}, and \emph{Estimation}. During the Selection phase, the algorithm selects the highest-ranked training example from the initial ICEs provided by the unsupervised retriever ranking method (out of $K$ ICEs). This selected example is then concatenated with the previously selected ICEs. In the first iteration, no ICEs have been selected before. %A predetermined template is used for ICE creation (refer to~\ref{prompt_creation}). 
In the Translation phase, the selected ICE is
%encoded using the LLM, and the resulting encoded text is then 
translated by the model. In the Estimation phase, the LLM output (translated text) and the original source text are inputted into the domain-specific QE model to estimate the quality of the translation. Our proposed methodology relies on sentence-level QE. 

Next, the selected ICE, together with its estimated quality and the LLM translation output, are appended to an intermediate list. To track the highest quality obtained thus far, the algorithm sorts the list in descending order based on the estimated quality. To avoid duplication, the selected ICE is removed before the next iteration. This iterative process continues until the best-estimated translation quality no longer improves within the specified patience threshold. Alternatively, the process terminates once all $K$ ICEs have been selected.

\begin{figure}[H]
    \centering
    \includegraphics[keepaspectratio,width=\textwidth]{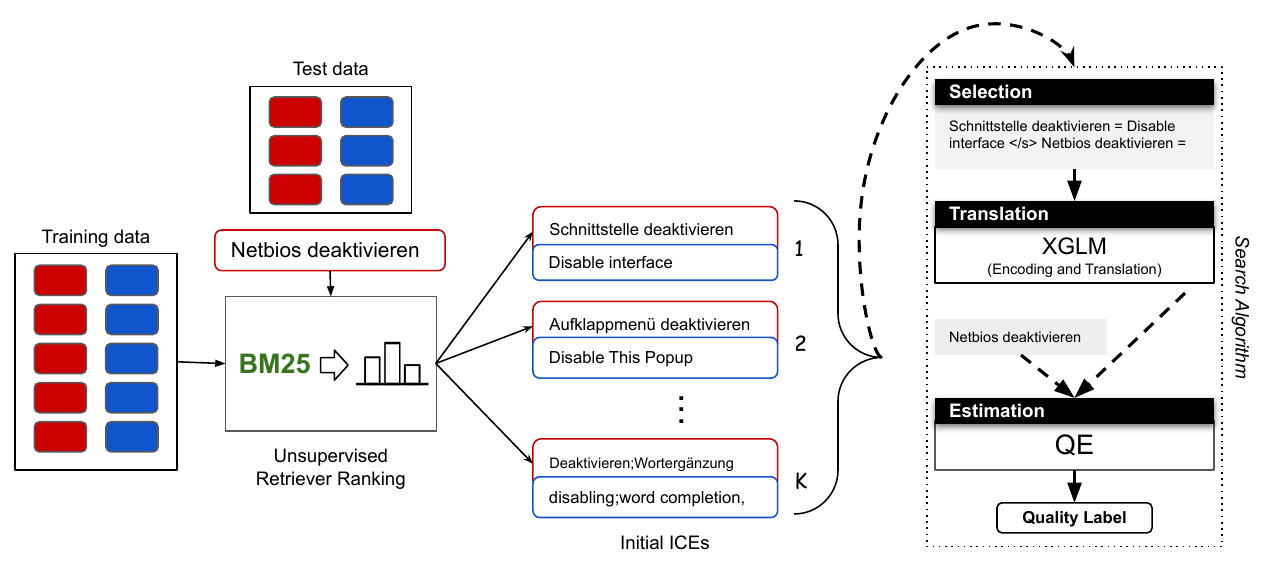}
\caption{\textbf{Overview illustration showing an iteration of our proposed methodology.} %``XGLM'' is the LLM %used in our experiments
%. ``QE'' represents the quality estimation model. %``Quality Label'' denotes the quality gold label, which in our experiment is BLEU.
}
    \label{fig:overview}
\end{figure}

This methodology allows for the systematic selection of ICEs that improve translation quality compared to previous ICL methodologies while efficiently managing the computational resources required for the search process. This efficiency is achieved by integrating early stopping conditions with predetermined patience. Notably, we do not explore permutations of initial ICEs, as doing so would require a large number of attempts, leading to high computational costs during the search process.\footnote{A pseudocode outlining the search methodology can be found in Algorithm~\ref{search_pseudo} in the Appendix. The phases of translating a source text of a test set using our methodology are depicted in Figure~\ref{fig:overview}.}

\section{Experiments Setup}
\label{sec:experiment_setup}
We conducted four main experiments to test the effectiveness of our methodology. Three of these experiments compare our methodology to existing ICL ones in different settings, or \emph{Modes}. %(detailed in Section~\ref{sec:modes}). %The fourth one compares our methodology to fine-tuning, aiming to assess which method is preferred (with respect to obtaining better translations).
The fourth experiment compares our methodology %as an ICL method 
to a fine-tuned mBART-50, aiming to assess which method is preferred (with respect to obtaining better translations).

It is important to note that we do not fine-tune the LLM. The process of building the QE model used in our experiments is detailed in Section~\ref{qe}. 

%Experiment 1 evaluates the effectiveness of our proposed methodology compared to existing methods in the literature. Experiment 2 explores the impact of re-ordering ICEs based on unigram overlaps versus using the standard BM25 ranking. Experiment 3 examines the reduction in BLEU score relative to Experiment 1. 

%\subsection{Unsupervised Retriever}\label{retrieval_module}
%We employed an unsupervised retriever method, BM25~\cite{10.1145/2682862.2682863}, to create the initial ICEs utilized by the proposed search algorithm. The initial number of ICEs was set at 16 to maintain consistency with earlier studies for a valid comparison.

%A detailed explanation of the implementation can be found in Section~\ref{sec:BM25}.
%implemented through \textit{BM25Okapi} in the \textit{rank\_bm25} package~\cite{10.1145/2682862.2682863}\footnote{\url{https://github.com/dorianbrown/rank_bm25}}. 

%\paragraph{Ordering ICEs Based on N-grams}  Given the use of ordering ICEs based on their n-gram overlap match with the source, as per~\cite{agrawal-etal-2023-context}, one of our objectives is to investigate how ordering the initial ICEs according to their n-grams (specifically unigrams) similarity with the source text affects the quality of translation. Therefore, in addition to ICEs ranked by BM25, we reorder them based on n-grams similarity. This involves tokenizing the sentences using NLTK word tokenizer and calculating the level of overlap between each. Sentences with greater levels of overlap matches will be given precedence in the list and fed into the LLMs earlier.

\subsection{Search Algorithm}
We conducted experiments using the search algorithm outlined in Section~\ref{sec:search_alg} across three operational modes:
% \subsubsection{Modes}\label{sec:modes}
% \begin{enumerate}[label=Mode \arabic*:, leftmargin=*]
\paragraph{Mode 1:} This mode uses QE with ICEs ordered by BM25 to assess the effectiveness of combining BM25 and QE in the proposed ICL methodology.
    
    %\item This mode entails the use of QE along with ICEs ordered by n-gram overlap, with the purpose of investigating the impact of n-gram ordering on the proposed methodology. 
\paragraph{Mode 2:} This mode investigates the impact of ordering ICEs by n-gram overlap, particularly unigrams, alongside QE, on the proposed methodology. Given the success of ordering ICEs based on their n-gram overlap match with the source, as demonstrated in~\citep{agrawal-etal-2023-context}, we assess how this ordering, based on ICEs' n-gram overlap with the source text, influences the translation quality. This involves reordering ICEs according to their n-gram overlap, which is calculated using the NLTK word tokenizer. Higher overlap matches prioritize ICEs in the list and feed them into LLMs earlier.
    
\paragraph{Mode 3:} %This mode does not involve QE and orders initial ICEs based on BM25 ranking. 
Instead of relying on QE, in this mode, we compute the BLUE score on the existing \textbf{test set}. This approach is not a realistic case, but it is the most favorable scenario, and we use it as the highest bound to compare with Mode 1.  
% \end{enumerate}

% \subsubsection{Early Stopping Conditions} 
The search algorithm generates up to 16 candidates. In each mode, we conducted experiments using three early stopping patience values (3, 8, and 16), determining the maximum number of ICEs ($K$) generated. We included Patience 16, which implies no early stopping, to evaluate the model's performance with the maximum ICEs. Additionally, the search process halts if the estimated label reaches or exceeds 100, preventing further evaluations.

%The search algorithm generates a maximum of 16 candidates. In each of the aforementioned modes, we conducted experiments using three distinct early stopping patience values, which set the maximum number of ICEs ($K$) that can be generated by the search algorithm. We experimented with values 3, 8, and 16, where Patience 16 implies no early stopping. Patience 16 was specifically examined in order to evaluate the model's performance with the maximum ICEs generated.

%In addition, we included a termination condition that activates when the (estimated) label reaches or exceeds 100. Upon activation, this condition halts the search process, preventing further attempts to evaluate other ICEs. %This feature has been implemented to enhance time effectiveness.

\subsection{Quality Estimation}\label{qe}
%Our proposed methodology requires a sentence-level QE model to effectively guide the selection process of the search algorithm. 
Following~\citep{ranasinghe-etal-2020-transquest,lee-2020-two,sharami-etal-2023-tailoring}, we develop a domain-specific QE model. %This development involves two key phases: 
First, we trained a QE model using out-of-domain (OOD) data (as detailed in Section~\ref{sec:syntheticQEdata}) to ensure generalizability; and second, we fine-tuned the model using the training set described in Section~\ref{sec:data} to provide domain-specific QE model and address domain mismatch, which is critical~\citep{koehn-knowles-2017-six}. %issue in both MT and QE

In our experiments, we used BLEU as the quality label because our study focused on translation performance rather than post-editing effort, which is typically evaluated using (H)TER~\citep{specia-farzindar-2010-estimating}. % in the context of QE experiments. 
We employed the ``MonoTransQuest'' architecture from the TransQuest framework~\citep{ranasinghe-etal-2020-transquest}, known for its success in prior QE studies. However, instead of employing softmax computation, we directly utilized logits to estimate the quality labels. This strategy saves computation time, as softmax computation can be resource-intensive~\citep{ruder2016wordembeddingspart2}. %but also eliminates the need to determine probabilities for each prediction in our experiments.

\subsubsection{QE data}\label{sec:syntheticQEdata}
We utilized the German-English ``EuroPat'' dataset, accessed through Opus~\citep{tiedemann-2012-parallel}%\footnote{\url{https://opus.nlpl.eu/}}
, to develop our generic QE model. We chose this dataset because it provides ample data samples, ensuring broad coverage of vocabulary –– a critical aspect in developing generic models.

However, as MT datasets like EuroPat typically consist of pairs of source and translated text, it was necessary to synthetically create post-editing text (since the QE data creation process requires a triplet input: source text, machine-translated text, and post-edited text). To accomplish this, we used a pre-trained multilingual MT model, namely mBART-50 that supported the language pair used in our experiment. This involves translating 1M randomly chosen source texts from EuroPat. Afterward, the resulting translations were considered as machine-translated text, with the corresponding reference translations acting as post-edited text.

Using SacreBLEU, we calculated the BLEU score, comparing the translated text with its corresponding post-edited text. This approach, which has been demonstrated to be effective in QE~\citep{negri-etal-2018-escape,lee-2020-two,sharami-etal-2023-tailoring}, enabled us to use the source and (machine-) translated text as input and the BLEU score as the target value for the QE model.
For building domain-specific QE, we utilized the training set detailed in Section~\ref{sec:data} and applied the aforementioned approach to synthetically generate BLEU scores for the entire dataset.

\subsection{Multilingual Large Language Model}\label{LLM}
For our experiments and hypothesis validation, we used XGLM~\citep{lin-etal-2022-shot}. %a language model developed by Facebook. 
This choice stems from the outstanding performance of the model in the MT field. This also ensures a fair comparison of our proposed methodology with previous research, such as~\citep{agrawal-etal-2023-context}, which introduced SOTA approaches in ICL for MT.

We used the 7.5 billion-parameter XGLM implementation and tokenizer by Hugging Face\footnote{\url{https://huggingface.co/docs/transformers/model_doc/xglm}}, consistent with previous research. % \subsubsection{Prompt Template}\label{prompt_creation}
%In our study, using the LLM detailed in Section~\ref{LLM}, 
We employed a template from~\cite{lin-etal-2022-shot} to maximize translation performance. $</s>$ serves as the ICE separator in this template.  ``BLANK'' denotes an empty string within the template.
\begin{equation}
{\small \begin{split}
\text{\{source text}_1\} & = \text{\{target text}_1\} </s> \\
\text{\{source text}_2\} & = \text{\{{target text}}_2\} </s> \\
\ldots  & = \ldots </s> \\
\text{\{source text}_n\} & = \text{BLANK} \notag
\end{split}}
\end{equation}

\subsection{Dataset and Evaluation Metrics}\label{sec:data}
We used a dataset comprising German-to-English translation pairs within the IT domain, sourced from~\citep{aharoni2020unsupervised}. This dataset was chosen due to the challenges that MT systems and LLMs face when translating out-of-domain contexts, particularly in specialized fields, as noted in previous studies~\citep{koehn-knowles-2017-six,agrawal-etal-2023-context}. The specialized and constrained nature of the IT domain provided an ideal setting for evaluating our methodology's performance under these conditions.

The dataset utilized in this study consisted of approximately 222k training sentences, 2k development sentences, and 2k test sentences. 
To assess the translation effectiveness of the models, we employed metrics such as BLEU from SacreBLEU~\citep{post-2018-call} and COMET~\citep{rei-etal-2020-comet}. 

\subsection{Number of ICEs}\label{sec:p_q}
We use between 1 and 16 ICEs. These may originate either from a random approach or from an advanced (guided) selection. To keep these separated in our analysis, we designate two different counts -- \( p \) and \( q \). This choice and naming convention is grounded by previous research exploring the impact of varying ICE numbers. While our study explicitly caps the upper limit of \( q \) at 16, values spanning from 1 to 16 remain feasible options –– unlike the fixed value in the compared systems.

\subsection{Compared Systems}\label{baselines}
%To evaluate the efficacy of our proposed ICL methodology in contrast to methodologies used in previous studies, we conducted a comparative analysis by comparing our findings to those obtained through different methods proposed in the existing literature. These methods consist of random sampling, task-level sampling, BM25, R-BM25, and mBART-50, which will be elaborated in the following paragraphs.
%To evaluate the efficacy of our proposed ICL methodology, 
We conducted a comparative analysis with methods from previous studies; \textit{random} and \textit{task-level sampling}, \textit{BM25}, \textit{R-BM25}, and \textit{fine-tuned mBART-50}. %Our baseline is defined as the method with the highest performance, detailed in Section~\ref{sec:SelectedBL}.
\paragraph{Random:} 
We conducted three random trials, generating random numbers based on parameter $p$. These numbers, ranging from 1 to the size of the training set, selected corresponding translation pairs. %In cases where multiple numbers are generated, we identify and choose the translation pairs associated with each of these numbers. 
%For example, if the randomly generated numbers are 1, 10, and 100, we select the respective examples from the training set. 
To create the prompt\footnote{In the literature, the term ``prompt'' is frequently used interchangeably with ``ICE''}, in addition to the training examples (i.e., ICEs), we need the source side intended for translation. We utilize the source from the development set, in contrast to the advanced methods in ICL, where the source text from the test set is typically employed. %That is, we concatenate the selected examples with each source text from the development set using the prompt template. %The reason for selecting the development set over the test set in this approach is that it has a higher potential to match the test set content.
The reason for selecting the development set over the test set in this approach is that development sets are generally from the same distribution, domain, and context as the test set. This similarity increases the likelihood that the examples in the development set will better match the content and context of the test set, thereby enhancing the relevance and effectiveness of the prompts.

The generated prompt is inputted into the LLM for translation. Then, the BLEU score of the development set is computed. The random number that produces the highest score among the trials is selected, and the training examples linked to this number are concatenated with the test set's source text.

\paragraph{Task-level:}
Based on the work of \citet{agrawal-etal-2023-context}, the task-level approach is similar to the random approach but differs in the number of trials used. We employ 100 trials for the task-level approach, a significantly higher number than the random approach. The reason for using more trials is to generate a greater variety of ICEs, aiming to enhance the performance of LLMs in the translation task. However, this results in longer execution times compared to the random approach.

\paragraph{BM25:}\label{sec:BM25}
%Previous studies suggest that unsupervised retrievers, such as BM25, offer substantial improvements in ICL performance compared to random and task-level approaches~\cite{liu-etal-2022-makes,luo2023dricl,wang2024learning}. 
Using the Moses Tokenizer~\citep{koehn-etal-2007-moses}, we first tokenize the training set's source samples. % based on their respective language. %\footnote{\url{https://github.com/luismsgomes/mosestokenizer}}
Then, a BM25 model is created for the tokenized corpus by employing the \textit{BM25Okapi} implementation within the \textit{rank\_bm25} package.%~\cite{10.1145/2682862.2682863}.
\footnote{\url{https://github.com/dorianbrown/rank_bm25}} 

Next, the test set is %iterated through, and each source text is 
tokenized using the tokenized source. 
The algorithm then searches for similar training samples based on BM25 criteria, selecting the top $q$ matches for the model. %We used the template mentioned earlier (refer to \ref{prompt_creation}) to create the prompts. 
This methodology utilizes the test set as opposed to random and task-level approaches using the development set.  

\paragraph{Re-rank BM25 (R-BM25):}
%BM25 aims to find translation examples with the highest n-gram overlap with the source sentence of the test set~\cite{luo2023dricl}. However, since there is no link between retrieved examples (i.e., they score independently), the top matches may not include all the n-grams present in the source text. This poses an issue in ICL since the input size of the LLMs is typically limited. That is, they are unable to receive a large number of ICEs to make up for the lack of coverage of certain n-grams. To address this issue, \cite{agrawal-etal-2023-context} proposed a re-rank version of BM25 called R-BM25.
BM25 aims to find translation examples with the highest n-gram overlap with the source sentence~\citep{luo2023dricl}. However, since retrieved examples score independently, top matches may lack coverage of all source n-grams. This poses an issue in ICL due to LLM input size limitations. To address this, \cite{agrawal-etal-2023-context} proposed R-BM25.
R-BM25 employs a recall-based n-gram overlap~\citep{agrawal-etal-2023-context} to extract word n-grams and their numbers from the test source and BM25 retrieved examples. %This is accomplished by computing a recall-based n-gram overlap score, which is explained in detail in the original paper (refer to~\cite{agrawal-etal-2023-context}). %The example that receives the highest score is included in the selected prompts set. For the next iteration of selection, the test source n-grams are down-weighted by a certain factor. The process is performed repeatedly until a specific threshold is reached.

\paragraph{Fine-tuning mBART-50:}
Different ICL methodologies, including our own, are assessed in comparison to the process of fine-tuning a pre-trained multilingual MT model, specifically mBART-50. The selection of mBART-50 is based on its alignment with the language specifications of the experiment and its proven track record of achieving success in MT tasks through the utilization of pre-trained models~\citep{Yuan2022AnIM,pham-etal-2022-effective}. The fine-tuning of mBART-50 is carried out using the training data outlined in Section~\ref{sec:data}.

% \vspace*{-15.35mm}
\subsection{Computational Costs}
We monitored and reported the computational costs of the models utilized in our experiments using the \emph{carbontracker} package.\footnote{\url{https://github.com/lfwa/carbontracker}} This involved calculating the carbon footprint (CO\textsubscript{2}eq) emissions, time to prediction (TTP), and electricity consumption (kWh) associated with our experiments. Our experiments were conducted using NVIDIA A40 GPUs.

The script for running our experiments is publicly available at \url{https://github.com/JoyeBright/ICLviaQE/}.

% \vspace*{-5.35mm}
\section{Experiments Results}\label{sec:results}
This section presents the results of our experiments. To ensure a fair comparison, we conducted a statistical analysis test (t-test) to determine if our models significantly outperformed the baseline. 

% \subsection{Compared Systems}\label{sec:SelectedBL}
Comparing to previous work, the results shown in Table~\ref{table:main-scores}, indicate that R-BM25 with 16 ICEs outperforms other methods. It is notable that there is a positive correlation between the number of examples and evaluation scores (consistent through all methods -- Random, Task-level, BM25, and R-BM25), although at the expense of prediction time (i.e., TTP). Employing 16 examples significantly improved performance compared to using only one example in the random approach. 
%The results from the methods in the literature, shown in Table~\ref{table:main-scores}, revealed that R-BM25 with 16 ICEs leads to superior performance compared to the other methods. As a result, we considered the R-BM25 model as our baseline for further comparative analysis. Moreover, while the random approach with only one ICE yielded a BLEU score of 10.38, indicating the lowest performance, this score significantly improved to 31.65 when employing 16 examples. This correlation between the number of examples and evaluation scores was observed across all methods (Random, Task-level, BM25, and R-BM25), although at the expense of prediction time (i.e., TTP).

% \subsection{Our Methodology}
%Table~\ref{table:main-scores} also shows the evaluation of our proposed methodology across the various modes. 
Analyzing the performance of our methods in Mode 1 (referred to as ``M 1'', with P = 3, 8, or 16 in Table~\ref{table:main-scores}), we observe that our proposed methodology with different patience thresholds consistently outperforms all previous methods, including the baseline. This trend holds for both the COMET and BLEU metrics across all the methods. Specifically, our method exhibits a minimum improvement of 0.52 points in the BLEU score (from 45.20 to 45.72) with patience threshold of 3 and a maximum improvement of 1.58 points in the BLEU score (from 45.20 to 46.78) with a patience threshold of 16 compared to R-BM25 with 16 examples.

\begin{table}[H]
\centering
\renewcommand{\arraystretch}{1.20}
\setlength\tabcolsep{1.5pt} 
\resizebox{\linewidth}{!}{%
\begin{tabular}{l||l|c|c|c|c|c}
\hline\hline
\textbf{Method}  & \textbf{\( p \) + \( q \)}    & \textbf{BLEU} & \textbf{COMET}  & \textbf{TTP}     & \textbf{CO2} & \textbf{GPU}\\
 & &  &  & \textbf{(hh:mm)}     & \textbf{(kg)} & \textbf{(kWh)} \\\hline
Random     & 1 + 0  & 10.38 & 0.6895 & 01:51 & 00.13       & 00.39      \\ 
Random     & 16 + 0 & 31.65 & 0.7844  & 02:20 & 00.19       & 00.58      \\ 
Task-level & 1 + 0  & 29.17 & 0.7586 & 62:50 & 09.83       & 29.10      \\ 
Task-level & 16 + 0 & 32.88 & 0.8083 & 78:30 & 12.80       & 35.91     \\ 
BM25              & 0 + 1  & 39.24 & 0.7833 & 00:56 & 00.06       & 00.19      \\ 
BM25              & 0 + 16 & 44.50 & 0.8120 & 00:58 & 00.07       & 00.19      \\ 
R-BM25            & 0 + 1  & 40.88 & 0.7990 & 01:01 & 00.06       & 00.21      \\ 
R-BM25            & 0 + 16 & \textbf{45.20} & \textbf{0.8218}  & 01:04 & 00.07       & 00.21      \\ \hline
M 1, P = 3     & 0 + 16 & 45.72 & 0.8395 & 01:49 & 00.22       & 00.67      \\ 
M 1, P = 8     & 0 + 16 & \textbf{46.43} & \textbf{0.8501}  & 03:48 & 00.50       & 01.51      \\ 
M 1, P = 16    & 0 + 16 & \textbf{46.78} & \textbf{0.8554}   & 05:11 & 00.68      & 02.05      \\ \hline
M 2, P = 3     & 0 + 16 & 46.05  & 0.8400  & 01:30 & 00.21       & 00.64      \\
M 2, P = 8     & 0 + 16 & \textbf{46.59}  & \textbf{0.8518}  & 03:52 & 00.51        & 01.52      \\
M 2, P = 16    & 0 + 16 & \textbf{46.52}  & \textbf{0.8564}  & 05:00 & 00.66       & 02.01      \\ \hline
M 3, P = 3     & 0 + 16 & 49.89  & 0.8532  & 01:36 & 00.22       & 00.66      \\ 
M 3, P = 8     & 0 + 16 & \textbf{52.63}  & \textbf{0.8725}  & 03:14 & 00.45       & 01.40      \\ 
M 3, P = 16    & 0 + 16 & \textbf{53.50}  &  \textbf{0.8791} & 04:08 & 00.55       & 01.65      \\ \hline
mBART-50          & N/A    & 42.76  & 0.8659 & 11:20 & 01.88      & 04.82 \\
\hline\hline
\end{tabular}
}
\caption{\textbf{Method Performance in BLEU and COMET Scores.} \( M \) 1 to 3 denotes Mode 1 to 3; \( P \) is the patience value; \( p \) and \( q \) are as defined in Section~\ref{sec:p_q}. ``N/A'' (not applicable) indicates that fine-tuning does not use ICEs. Bold font represents the highest translation performance. Two numbers are in bold if they are statistically similar (t-test, \( p\_value = 0.05 \)).}
\label{table:main-scores}
\end{table}

\vspace*{-2.35mm}

Consequently, our methods in Mode 1 are ranked based on their performance, with patience 3 being the least effective model, followed by patience 8, and finally patience 16, representing the most effective method. This ranking indicates that increasing the patience threshold can significantly enhance the translation performance. However, the improvement with patience 16 is not statistically significant compared to patience 8, suggesting that more ICEs do not necessarily enhance translation performance. Similarly, while more substantial contextual improvement (as indicated by the COMET) is observed at the maximum patience threshold (16), it is not statistically significant compared to patience 8.

The Mode 2 results demonstrate that all three patience thresholds surpass the methods in the literature. However, this improvement is not statistically significant when compared with the respective experiments in Mode 1. This suggests that ordering the examples according to n-gram (unigram) similarity does not enhance the translation performance in our methodology.

When it comes to Mode 3, we should stress that this is an unrealistic scenario, but used as the highest bound. The results indicate that with a patience of 3, the BLEU score is 4.17 points lower (49.89-45.72). With a patience of 8, this gap increases to 6.2 points (52.63-46.43), and with a patience of 16, it widens further to 6.72 points (53.50-46.78). These differences arise from the QE model estimations in our experiment compared to the scenario where reference labels are available to the search algorithm.

\subsection{Time to Prediction (TTP)}
Among the methods examined, task-level execution required the most time, with approximately 62 hours for one example and 78 hours for 16 examples. Our method (Mode 1) with a patience value of 16 is relatively time-intensive, taking approximately 5 hours, while a patience value of 3 is comparable to the baseline method, differing by only around 50 minutes. Mode 2 is nearly equivalent to Mode 1 in terms of TTP, whereas Mode 3, where the reference labels are accessed, requires less time than Modes 1 and 2. %This is primarily because the search algorithm accesses the reference labels. 
In addition, the search algorithm incorporates a termination condition, and given that QE estimation rarely triggers this condition, numerous ICEs are left unattempted, resulting in significant time savings. 

It is also important to note the time required to train the QE models used in the prediction process. As provided in Appendix~\ref{app:QEstats}, the training time for the generic QE model is $+/-$ 5 hours and 55 minutes, while the specific QE model takes about $+/-$~6 hours and 54 minutes. Although these training times are significant, it is crucial to recognize that QE models, similar to MT models, can be reused for the same language pair and domain, thereby amortizing the initial training cost over multiple predictions.

% \subsection{Fine-tuning vs. ICL}
The last row of Table~\ref{table:main-scores} shows the scores of the translations obtained with the mBART-50 model fine-tuned on the same training set as in ICL. %results concerning the fine-tuning of a pre-trained multilingual model, specifically mBART-50, using the same training dataset as in ICL (referred to as ``FT: mBART-50''). 
Despite mBART-50 being tailored for MT across 50 languages, it did not outperform the R-BM25 method with 16 examples (best from the existing methods); it was better only than Random, Task-level, BM25, and R-BM25, each with only 1 example. However, when considering translation performance from a contextual perspective, the COMET results indicate that fine-tuning mBART-50 leads to superior performance compared with lexical overlap. Nevertheless, fine-tuning took significantly longer than identifying ICEs and obtaining inferences from the XGLM.

Compared to our methodology, especially when considering the least performing method (M 1, P = 3), it is significantly worse -- 6.47\% (42.76 to 45.72). This highlights the substantial efficacy of ICL compared to fine-tuning. Nonetheless, it is noteworthy that various factors might contribute to this observation: e.g., the model's size might be a critical factor, especially during deployment, where larger models like XGLM could pose challenges.

\section{Analysis}\label{sec:analysis}
\begin{figure}[H]
    \centering
    \includegraphics[width=0.85\linewidth]{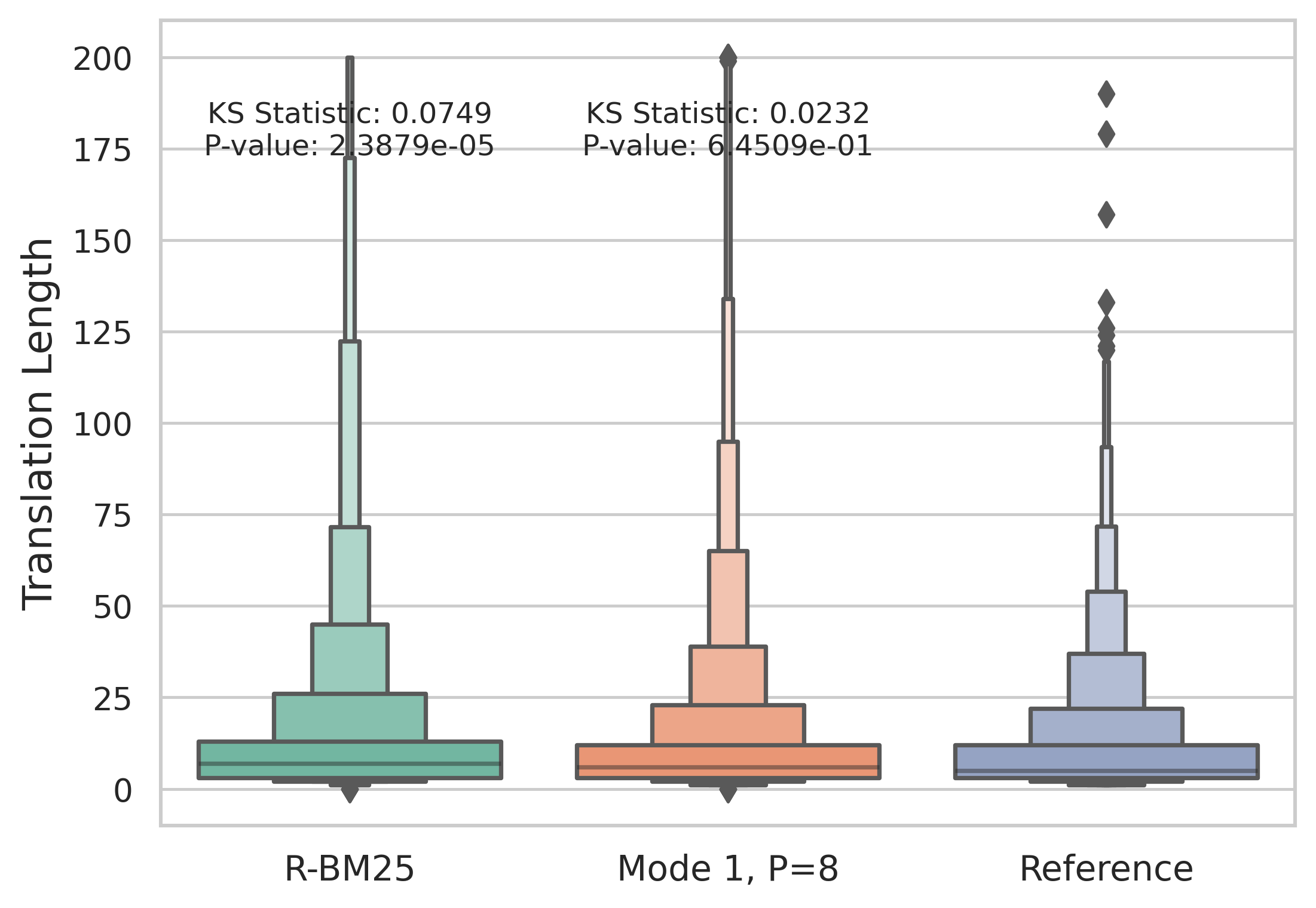} 
    \caption{\textbf{Tokenized Translation Lengths} comparison between R-BM25, our Mode 1, P=8, and the reference. ``KS'' denotes the Kolmogorov-Smirnov test, with the p-value indicating significance.}
    \label{fig:length_output}
\end{figure}

\paragraph{Output analysis} Pre-trained LLMs often exhibit over-generation, i.e., the generation of a larger number of tokens than expected by a human (in comparison to a reference), necessitating extensive post-processing (e.g., post-editing)~% to refine their output for conciseness and relevance~
\citep{bawden-yvon-2023-investigating}. Figure~\ref{fig:length_output} shows the tokenized output lengths (translations) for our model (Mode 1, patience 8),\footnote{Our other models in Mode 1 exhibited similar distributions.} alongside the R-BM25 with 16 examples. The analysis shows that the length distributions for both models align with the reference distribution, suggesting that the models do not over-generate. 

To quantitatively compare these distributions to the reference, we employed the Kolmogorov-Smirnov~(KS) test~\citep{kolmogorov1933sulla}. The results indicate that for R-BM25 versus the reference, the KS statistic is relatively high (\(0.0749\)), reflecting a significant difference between the translation lengths of R-BM25 and the reference distribution. The extremely low p-value (\(2.39 \times 10^{-5}\)) further confirms this significant discrepancy. Conversely, for Mode 1 with P=8 versus the reference, the KS statistic is considerably lower (\(0.0232\)), indicating a much smaller difference in translation lengths. The higher p-value (\(0.6451\)) suggests no significant difference, implying that the distribution of Mode 1, P=8 is similar to the reference distribution. 

These findings suggest that our proposed methodology could yield translations closer in length to the reference, potentially reducing the need for labor-intensive post-processing efforts and enhancing computational efficiency.
%but also streamlines the overall translation process.
% \vspace*{-2.00mm}

\paragraph{ICE Number Analysis}
The number of selected ICEs holds a significant importance within the ICL algorithm, as it directly impacts the token processing time and the capacity of LLMs to handle additional ICE instances. % (due to their limited capacity~\cite{agrawal-etal-2023-context}). 
We analyzed the number of ICEs that our algorithm selected %(where a pair of source and target counts as one ICE) 
across all three modes. The results (Table~\ref{table:ICE_Count}) show that the minimum number of ICEs selected is 1, while the maximum is: 12 for Mode 1, 16 for Mode 2, and 16 for Mode 3.
The average (mean) number of ICEs is found to be lowest in Mode 3 and highest in Mode 1. In addition, Mode 2 results in a reduction in the number of ICEs within our proposed algorithm. The notably lower average number of ICE instances in Mode 3 can be attributed to its access to the test set, allowing for the selection of optimal ICE combinations based on test set performance and activating an early stopping condition if the score exceeds 100. Contrarily, while Mode 1 exhibits similarities to Mode 3, its relatively higher average can be linked to inaccuracies in QE estimation. Moreover, our analysis shows that QE estimations rarely reach a score of 100, thus rendering the early stopping condition inactive. %for that aspect.

\begin{table}[H]
\centering
\resizebox{\linewidth}{!}{%
\begin{tabular}{l|c|c|c}
\hline\hline
 Mode   & Min                      & Mean                     & Max                         \\
\hline
 \#1 & [1, 1, 1] & [2.25, 3.76, 4.84] & [12, 16, 16] \\
 \#2 & [1, 1, 1] & [2.20, 3.70, 4.74] & [12, 16, 16] \\
 \#3 & [1, 1, 1] & [2.15, 3.47, 4.47] & [12, 16, 16] \\
\hline\hline
\end{tabular}
}
\caption{\textbf{Number of ICEs selected for each mode at different patience thresholds.} Labels [x, y, z] correspond to patience values 3, 8, and 16.}
\label{table:ICE_Count}
\end{table}

\paragraph{CO\textsubscript{2} Emissions}
%Besides evaluating the translation performance of the models, it is crucial to consider their computational costs, notably the associated CO\textsubscript{2} emissions. 
%Considering CO\textsubscript{2} emissions alongside translation performance is crucial.  
Our analysis reveals that using XGLM for translation yields lower CO\textsubscript{2} emissions than fine-tuning mBART-50, making it a more environmentally sustainable choice. In Mode 1 of our proposed methodology, with patience 16, XGLM emitted 0.68 KG of CO\textsubscript{2}, while fine-tuning mBART-50 emitted 1.88 KG. Interestingly, the task-level method with 16 ICEs emitted the highest amount of CO\textsubscript{2}, totaling 12.80 KG. Our proposed approach leads to higher CO\textsubscript{2} emissions than R-BM25.

\section{Related Work}\label{sec:related_work}
\paragraph{ICL for MT.} ICL\footnote{Also referred to as the prompt retrieval method} represents a relatively new paradigm in natural language understanding. Unlike traditional fine-tuning approaches, where a PLM undergoes parameter updates using a specific dataset, ICL typically directly generates the output without any modification to its parameters \citep{Radford2019LanguageMA,NEURIPS2020_1457c0d6}. This is achieved by solely providing the model with a few examples, known as ICEs, which prime the PLM to enhance its performance for the given task~\citep{10.1162/tacl_a_00324}. %The primary role of ICEs is to prime the PLM, enhancing its performance to generate more favorable results in accordance with the given task \cite{10.1162/tacl_a_00324}.

As shown by~\cite{vilar-etal-2023-prompting}, the quality of translation is directly proportionate to the quality of ICEs, where quality refers to ICEs being relevant, clear, accurate, and domain-specific. However, considering all ICEs during processing is computationally demanding~\citep{alves-etal-2023-steering}. Hence, it is crucial to selectively choose ICEs that can enhance MT quality. %, particularly considering the limited input capacity of LLMs. Various studies have explored different strategies in MT to address these challenges. 
\cite{goyal-etal-2022-flores} conducted a study where ICEs were randomly selected. Despite finding that this random selection of ICEs resulted in good translation performance, the neglect of their order, which was identified as important~\citep{liu-etal-2022-makes,lu-etal-2022-fantastically}, was a drawback in this approach. To address this, methodologies such as~\citep{agrawal-etal-2023-context} introduced a re-ranking technique (R-BM25). However, their methodology relies solely on n-grams to order examples, which can enhance fluency but may overlook contextual factors. In our approach, we investigated the unigram order of initial ICEs provided by the BM25 algorithm. We leave the in-depth analysis of ICE order for future work. Additionally, \citet{m-etal-2023-ctqscorer} highlighted the advantages of using multiple features in ICE selection to improve translation quality, while our QE-based approach simplifies ICE selection without needing to generate additional features, ensuring efficiency.

\paragraph{QE in MT Evaluation.} QE models offer a quick solution to the assessment of the overall usefulness of translated text. %, typically produced by an MT system.\footnote{Given that a QE system takes as input the source and its translation, it is not strictly required that the translation is generated by an MT system; however this is the standard use-case of QE.} 
These models do not rely on reference translations, thereby reducing the human effort required for quality evaluation~\citep{tamchyna-2021-deploying, murgolo-etal-2022-quality, zerva-etal-2022-findings,blain-etal-2023-findings}.
Similar to MT models, previous studies highlight the importance of domain-specific QE for accurately estimating translation quality across diverse domains~\citep{lee-2020-two, sharami-etal-2023-tailoring}. This is why, in our work, we employed a domain-specific QE model instead of a generic one to enhance the selection of ICEs. %By doing so, we aimed to improve the quality estimation of domain-specific translations, thereby enhancing overall translation performance.

Integrating QE into ICL offers significant, yet largely unexplored, potential. %QE can quantify the fluency and adequacy of translations without needing reference translations. 
QE can also better capture out-of-domain gender and word-sense-disambiguation errors~\citep{dinh-niehues-2023-perturbation}. Additionally, integrating QE can mitigate reference bias, a significant challenge in accurately estimating the output quality of LLMs%, particularly in sequence transduction tasks 
~\citep{goyal2023news,raunak-etal-2023-gpts}. The introduction of COMET-QE \citep{raunak-etal-2023-dissecting} exemplifies this pursuit, providing a metric tailored to evaluate the quality of perturbed prompts provided to GPT-3 \citep{NEURIPS2020_1457c0d6}, aiming to mitigate reference bias. While in our approach, we employ domain-specific QE to guide the selection of ICEs, this underscores the potential of QE in refining LLM inputs (i.e., ICEs). %Therefore, leveraging QE within ICL has the potential to streamline ICE selection, thereby ensuring improved translation quality and efficiency in LLM-based translation systems.
% \vspace*{-2.0mm}
\section{Conclusion}\label{sec:conclusion}%\vspace*{-1.0mm}
We propose a novel in-context learning (ICL) methodology for enhancing the translation capabilities of large language models (LLMs) while optimizing computational resources. Our approach leverages domain-specific quality estimation (QE) to guide in-context selection, particularly focusing on determining the suboptimal number and the combinations of in-context examples (ICEs). This novel strategy moves beyond the conventional reliance solely on translation references from development sets seen in prior methods. %Specifically, our methodology comprises two main phases: (1) employing an unsupervised retriever, namely BM25, to identify relevant examples, and (2) utilizing a search algorithm guided by the QE model to select a combination of examples that enhance translation quality.

We evaluated our approach across different modes and early stopping patience values on the German-to-English IT dataset. Our experiments consistently showed the superior performance of our methodology, surpassing all prior works across both BLEU and COMET metrics. Our method consistently improves BLEU scores, although this comes at the cost of increased computation time. We also investigated the impact of ordering the ICEs based on their unigram overlap with the source text and found it to be not statistically significant. Furthermore, our experiments highlighted the value of ICL compared to fine-tuning a pre-trained large model, namely mBART-50. We also highlighted that our method leads to less carbon emissions while achieving better translation performance.
%, gaining insights into its performance across various scenarios.
%, including the selected baseline model, 

In the future, we would like to conduct further research on the impact of our proposed methodology across different language pairs, domains and LLMs. Also, we aim to explore alternative metrics beyond BLEU to tailor the selection process, as well as additional features such as bigram, type/token ratio, and length when ordering examples prior to their input into LLMs. %Additionally, a comparison between the fine-tuning of the XGLM (using the parallel dataset) and the proposed ICL methodology could be conducted.  
% \vspace*{4.0mm}
%Several avenues for future research stem from our methodology and findings. Firstly, assessing the generalizability of our methodology by testing it on additional datasets and LLMs could provide valuable insights. Secondly, exploring alternative metrics beyond BLEU to tailor the selection process based on specific quality labels other than BLEU could be beneficial. This also extends to Mode 2, where incorporating additional features (such as bigram, type/token ratio, and length) to prioritize the order of ICEs is feasible. Lastly, a comparison between the fine-tuning of the XGLM (using the parallel dataset) and the proposed methodology could be conducted.

\section{Acknowledgments}
We would like to thank the anonymous reviewers for their helpful comments.

\begin{small}
\bibliographystyle{apalike}
\bibliography{amta2024}

\begin{thebibliography}{}

\bibitem[Agrawal et~al., 2023]{agrawal-etal-2023-context}
Agrawal, S., Zhou, C., Lewis, M., Zettlemoyer, L., and Ghazvininejad, M. (2023).
\newblock In-context examples selection for machine translation.
\newblock In Rogers, A., Boyd-Graber, J., and Okazaki, N., editors, {\em Findings of the Association for Computational Linguistics: ACL 2023}, pages 8857--8873, Toronto, Canada. Association for Computational Linguistics.

\bibitem[Aharoni and Goldberg, 2020]{aharoni2020unsupervised}
Aharoni, R. and Goldberg, Y. (2020).
\newblock Unsupervised domain clusters in pretrained language models.
\newblock In {\em Proceedings of the 58th Annual Meeting of the Association for Computational Linguistics (Volume 1: Long Papers)}. Association for Computational Linguistics.

\bibitem[Alves et~al., 2023]{alves-etal-2023-steering}
Alves, D., Guerreiro, N., Alves, J., Pombal, J., Rei, R., de~Souza, J., Colombo, P., and Martins, A. (2023).
\newblock Steering large language models for machine translation with finetuning and in-context learning.
\newblock In Bouamor, H., Pino, J., and Bali, K., editors, {\em Findings of the Association for Computational Linguistics: EMNLP 2023}, pages 11127--11148, Singapore. Association for Computational Linguistics.

\bibitem[Bawden and Yvon, 2023]{bawden-yvon-2023-investigating}
Bawden, R. and Yvon, F. (2023).
\newblock Investigating the translation performance of a large multilingual language model: the case of {BLOOM}.
\newblock In Nurminen, M., Brenner, J., Koponen, M., Latomaa, S., Mikhailov, M., Schierl, F., Ranasinghe, T., Vanmassenhove, E., Vidal, S.~A., Aranberri, N., Nunziatini, M., Escart{\'\i}n, C.~P., Forcada, M., Popovic, M., Scarton, C., and Moniz, H., editors, {\em Proceedings of the 24th Annual Conference of the European Association for Machine Translation}, pages 157--170, Tampere, Finland. European Association for Machine Translation.

\bibitem[Blain et~al., 2023]{blain-etal-2023-findings}
Blain, F., Zerva, C., Rei, R., Guerreiro, N.~M., Kanojia, D., C.~de Souza, J.~G., Silva, B., Vaz, T., Jingxuan, Y., Azadi, F., Orasan, C., and Martins, A. (2023).
\newblock Findings of the {WMT} 2023 shared task on quality estimation.
\newblock In Koehn, P., Haddow, B., Kocmi, T., and Monz, C., editors, {\em Proceedings of the Eighth Conference on Machine Translation}, pages 629--653, Singapore. Association for Computational Linguistics.

\bibitem[Brown et~al., 2020]{NEURIPS2020_1457c0d6}
Brown, T., Mann, B., Ryder, N., Subbiah, M., Kaplan, J.~D., Dhariwal, P., Neelakantan, A., Shyam, P., Sastry, G., Askell, A., Agarwal, S., Herbert-Voss, A., Krueger, G., Henighan, T., Child, R., Ramesh, A., Ziegler, D., Wu, J., Winter, C., Hesse, C., Chen, M., Sigler, E., Litwin, M., Gray, S., Chess, B., Clark, J., Berner, C., McCandlish, S., Radford, A., Sutskever, I., and Amodei, D. (2020).
\newblock Language models are few-shot learners.
\newblock In Larochelle, H., Ranzato, M., Hadsell, R., Balcan, M., and Lin, H., editors, {\em Advances in Neural Information Processing Systems}, volume~33, pages 1877--1901. Curran Associates, Inc.

\bibitem[Dinh and Niehues, 2023]{dinh-niehues-2023-perturbation}
Dinh, T.~A. and Niehues, J. (2023).
\newblock Perturbation-based {QE}: An explainable, unsupervised word-level quality estimation method for blackbox machine translation.
\newblock In Utiyama, M. and Wang, R., editors, {\em Proceedings of Machine Translation Summit XIX, Vol. 1: Research Track}, pages 59--71, Macau SAR, China. Asia-Pacific Association for Machine Translation.

\bibitem[Goyal et~al., 2022]{goyal-etal-2022-flores}
Goyal, N., Gao, C., Chaudhary, V., Chen, P.-J., Wenzek, G., Ju, D., Krishnan, S., Ranzato, M., Guzm{\'a}n, F., and Fan, A. (2022).
\newblock The {F}lores-101 evaluation benchmark for low-resource and multilingual machine translation.
\newblock {\em Transactions of the Association for Computational Linguistics}, 10:522--538.

\bibitem[Goyal et~al., 2023]{goyal2023news}
Goyal, T., Li, J.~J., and Durrett, G. (2023).
\newblock News summarization and evaluation in the era of gpt-3.

\bibitem[Jiang et~al., 2020]{10.1162/tacl_a_00324}
Jiang, Z., Xu, F.~F., Araki, J., and Neubig, G. (2020).
\newblock {How Can We Know What Language Models Know?}
\newblock {\em Transactions of the Association for Computational Linguistics}, 8:423--438.

\bibitem[Koehn et~al., 2007]{koehn-etal-2007-moses}
Koehn, P., Hoang, H., Birch, A., Callison-Burch, C., Federico, M., Bertoldi, N., Cowan, B., Shen, W., Moran, C., Zens, R., Dyer, C., Bojar, O., Constantin, A., and Herbst, E. (2007).
\newblock {M}oses: Open source toolkit for statistical machine translation.
\newblock In Ananiadou, S., editor, {\em Proceedings of the 45th Annual Meeting of the Association for Computational Linguistics Companion Volume Proceedings of the Demo and Poster Sessions}, pages 177--180, Prague, Czech Republic. Association for Computational Linguistics.

\bibitem[Koehn and Knowles, 2017]{koehn-knowles-2017-six}
Koehn, P. and Knowles, R. (2017).
\newblock Six challenges for neural machine translation.
\newblock In Luong, T., Birch, A., Neubig, G., and Finch, A., editors, {\em Proceedings of the First Workshop on Neural Machine Translation}, pages 28--39, Vancouver. Association for Computational Linguistics.

\bibitem[Kolmogorov, 1933]{kolmogorov1933sulla}
Kolmogorov, A.~N. (1933).
\newblock Sulla determinazione empirica di una legge di distribuzione.
\newblock {\em Giornale dell'Istituto Italiano degli Attuari}, 4:83--91.

\bibitem[Kumar et~al., 2023]{m-etal-2023-ctqscorer}
Kumar, A., Puduppully, R., Dabre, R., and Kunchukuttan, A. (2023).
\newblock {CTQS}corer: Combining multiple features for in-context example selection for machine translation.
\newblock In Bouamor, H., Pino, J., and Bali, K., editors, {\em Findings of the Association for Computational Linguistics: EMNLP 2023}, pages 7736--7752, Singapore. Association for Computational Linguistics.

\bibitem[Lee, 2020]{lee-2020-two}
Lee, D. (2020).
\newblock Two-phase cross-lingual language model fine-tuning for machine translation quality estimation.
\newblock In Barrault, L., Bojar, O., Bougares, F., Chatterjee, R., Costa-juss{\`a}, M.~R., Federmann, C., Fishel, M., Fraser, A., Graham, Y., Guzman, P., Haddow, B., Huck, M., Yepes, A.~J., Koehn, P., Martins, A., Morishita, M., Monz, C., Nagata, M., Nakazawa, T., and Negri, M., editors, {\em Proceedings of the Fifth Conference on Machine Translation}, pages 1024--1028, Online. Association for Computational Linguistics.

\bibitem[Lin et~al., 2022]{lin-etal-2022-shot}
Lin, X.~V., Mihaylov, T., Artetxe, M., Wang, T., Chen, S., Simig, D., Ott, M., Goyal, N., Bhosale, S., Du, J., Pasunuru, R., Shleifer, S., Koura, P.~S., Chaudhary, V., O{'}Horo, B., Wang, J., Zettlemoyer, L., Kozareva, Z., Diab, M., Stoyanov, V., and Li, X. (2022).
\newblock Few-shot learning with multilingual generative language models.
\newblock In Goldberg, Y., Kozareva, Z., and Zhang, Y., editors, {\em Proceedings of the 2022 Conference on Empirical Methods in Natural Language Processing}, pages 9019--9052, Abu Dhabi, United Arab Emirates. Association for Computational Linguistics.

\bibitem[Liu et~al., 2022]{liu-etal-2022-makes}
Liu, J., Shen, D., Zhang, Y., Dolan, B., Carin, L., and Chen, W. (2022).
\newblock What makes good in-context examples for {GPT}-3?
\newblock In Agirre, E., Apidianaki, M., and Vuli{\'c}, I., editors, {\em Proceedings of Deep Learning Inside Out (DeeLIO 2022): The 3rd Workshop on Knowledge Extraction and Integration for Deep Learning Architectures}, pages 100--114, Dublin, Ireland and Online. Association for Computational Linguistics.

\bibitem[Lu et~al., 2022]{lu-etal-2022-fantastically}
Lu, Y., Bartolo, M., Moore, A., Riedel, S., and Stenetorp, P. (2022).
\newblock Fantastically ordered prompts and where to find them: Overcoming few-shot prompt order sensitivity.
\newblock In Muresan, S., Nakov, P., and Villavicencio, A., editors, {\em Proceedings of the 60th Annual Meeting of the Association for Computational Linguistics (Volume 1: Long Papers)}, pages 8086--8098, Dublin, Ireland. Association for Computational Linguistics.

\bibitem[Luo et~al., 2023]{luo2023dricl}
Luo, M., Xu, X., Dai, Z., Pasupat, P., Kazemi, M., Baral, C., Imbrasaite, V., and Zhao, V.~Y. (2023).
\newblock Dr.icl: Demonstration-retrieved in-context learning.

\bibitem[Murgolo et~al., 2022]{murgolo-etal-2022-quality}
Murgolo, E., Sharami, J. P.~R., and Shterionov, D. (2022).
\newblock A quality estimation and quality evaluation tool for the translation industry.
\newblock In {\em Proceedings of the 23rd Annual Conference of the European Association for Machine Translation}, pages 307--308, Ghent, Belgium. European Association for Machine Translation.

\bibitem[Negri et~al., 2018]{negri-etal-2018-escape}
Negri, M., Turchi, M., Chatterjee, R., and Bertoldi, N. (2018).
\newblock {ESCAPE}: a large-scale synthetic corpus for automatic post-editing.
\newblock In {\em Proceedings of the Eleventh International Conference on Language Resources and Evaluation ({LREC} 2018)}, Miyazaki, Japan. European Language Resources Association (ELRA).

\bibitem[Petrov et~al., 2023]{petrov2023language}
Petrov, A., Malfa, E.~L., Torr, P. H.~S., and Bibi, A. (2023).
\newblock Language model tokenizers introduce unfairness between languages.

\bibitem[Pham et~al., 2022]{pham-etal-2022-effective}
Pham, N.-Q., Nguyen, T.~N., Nguyen, T.-B., Liu, D., Mullov, C., Niehues, J., and Waibel, A. (2022).
\newblock Effective combination of pretrained models - {KIT}@{IWSLT}2022.
\newblock In Salesky, E., Federico, M., and Costa-juss{\`a}, M., editors, {\em Proceedings of the 19th International Conference on Spoken Language Translation (IWSLT 2022)}, pages 190--197, Dublin, Ireland (in-person and online). Association for Computational Linguistics.

\bibitem[Post, 2018]{post-2018-call}
Post, M. (2018).
\newblock A call for clarity in reporting {BLEU} scores.
\newblock In {\em Proceedings of the Third Conference on Machine Translation: Research Papers}, pages 186--191, Belgium, Brussels. Association for Computational Linguistics.

\bibitem[Radford et~al., 2019]{Radford2019LanguageMA}
Radford, A., Wu, J., Child, R., Luan, D., Amodei, D., and Sutskever, I. (2019).
\newblock Language models are unsupervised multitask learners.

\bibitem[Ranasinghe et~al., 2020]{ranasinghe-etal-2020-transquest}
Ranasinghe, T., Orasan, C., and Mitkov, R. (2020).
\newblock {T}rans{Q}uest: Translation quality estimation with cross-lingual transformers.
\newblock In Scott, D., Bel, N., and Zong, C., editors, {\em Proceedings of the 28th International Conference on Computational Linguistics}, pages 5070--5081, Barcelona, Spain (Online). International Committee on Computational Linguistics.

\bibitem[Raunak et~al., 2023a]{raunak-etal-2023-dissecting}
Raunak, V., Menezes, A., and Awadalla, H. (2023a).
\newblock Dissecting in-context learning of translations in {GPT}-3.
\newblock In Bouamor, H., Pino, J., and Bali, K., editors, {\em Findings of the Association for Computational Linguistics: EMNLP 2023}, pages 866--872, Singapore. Association for Computational Linguistics.

\bibitem[Raunak et~al., 2023b]{raunak-etal-2023-gpts}
Raunak, V., Menezes, A., Post, M., and Hassan, H. (2023b).
\newblock Do {GPT}s produce less literal translations?
\newblock In Rogers, A., Boyd-Graber, J., and Okazaki, N., editors, {\em Proceedings of the 61st Annual Meeting of the Association for Computational Linguistics (Volume 2: Short Papers)}, pages 1041--1050, Toronto, Canada. Association for Computational Linguistics.

\bibitem[Rei et~al., 2020]{rei-etal-2020-comet}
Rei, R., Stewart, C., Farinha, A.~C., and Lavie, A. (2020).
\newblock {COMET}: A neural framework for {MT} evaluation.
\newblock In Webber, B., Cohn, T., He, Y., and Liu, Y., editors, {\em Proceedings of the 2020 Conference on Empirical Methods in Natural Language Processing (EMNLP)}, pages 2685--2702, Online. Association for Computational Linguistics.

\bibitem[Robertson and Zaragoza, 2009]{10.1561/1500000019}
Robertson, S. and Zaragoza, H. (2009).
\newblock The probabilistic relevance framework: Bm25 and beyond.
\newblock {\em Found. Trends Inf. Retr.}, 3(4):333–389.

\bibitem[Ruder, 2016]{ruder2016wordembeddingspart2}
Ruder, S. (2016).
\newblock {On word embeddings - Part 2: Approximating the Softmax}.
\newblock \url{http://ruder.io/word-embeddings-softmax}.

\bibitem[Sharami et~al., 2023]{sharami-etal-2023-tailoring}
Sharami, J. P.~R., Shterionov, D., Blain, F., Vanmassenhove, E., Sisto, M.~D., Emmery, C., and Spronck, P. (2023).
\newblock Tailoring domain adaptation for machine translation quality estimation.
\newblock In Nurminen, M., Brenner, J., Koponen, M., Latomaa, S., Mikhailov, M., Schierl, F., Ranasinghe, T., Vanmassenhove, E., Vidal, S.~A., Aranberri, N., Nunziatini, M., Escart{\'\i}n, C.~P., Forcada, M., Popovic, M., Scarton, C., and Moniz, H., editors, {\em Proceedings of the 24th Annual Conference of the European Association for Machine Translation}, pages 9--20, Tampere, Finland. European Association for Machine Translation.

\bibitem[Sia and Duh, 2023]{sia-duh-2023-context}
Sia, S. and Duh, K. (2023).
\newblock In-context learning as maintaining coherency: A study of on-the-fly machine translation using large language models.
\newblock In Utiyama, M. and Wang, R., editors, {\em Proceedings of Machine Translation Summit XIX, Vol. 1: Research Track}, pages 173--185, Macau SAR, China. Asia-Pacific Association for Machine Translation.

\bibitem[Specia and Farzindar, 2010]{specia-farzindar-2010-estimating}
Specia, L. and Farzindar, A. (2010).
\newblock Estimating machine translation post-editing effort with {HTER}.
\newblock In Zhechev, V., editor, {\em Proceedings of the Second Joint EM+/CNGL Workshop: Bringing MT to the User: Research on Integrating MT in the Translation Industry}, pages 33--43, Denver, Colorado, USA. Association for Machine Translation in the Americas.

\bibitem[Tamchyna, 2021]{tamchyna-2021-deploying}
Tamchyna, A. (2021).
\newblock Deploying {MT} quality estimation on a large scale: Lessons learned and open questions.
\newblock In {\em Proceedings of Machine Translation Summit XVIII: Users and Providers Track}, pages 291--305, Virtual. Association for Machine Translation in the Americas.

\bibitem[Tang et~al., 2020]{tang2020multilingual}
Tang, Y., Tran, C., Li, X., Chen, P.-J., Goyal, N., Chaudhary, V., Gu, J., and Fan, A. (2020).
\newblock Multilingual translation with extensible multilingual pretraining and finetuning.

\bibitem[Tiedemann, 2012]{tiedemann-2012-parallel}
Tiedemann, J. (2012).
\newblock Parallel data, tools and interfaces in {OPUS}.
\newblock In Calzolari, N., Choukri, K., Declerck, T., Do{\u{g}}an, M.~U., Maegaard, B., Mariani, J., Moreno, A., Odijk, J., and Piperidis, S., editors, {\em Proceedings of the Eighth International Conference on Language Resources and Evaluation ({LREC}'12)}, pages 2214--2218, Istanbul, Turkey. European Language Resources Association (ELRA).

\bibitem[Trotman et~al., 2014]{10.1145/2682862.2682863}
Trotman, A., Puurula, A., and Burgess, B. (2014).
\newblock Improvements to bm25 and language models examined.
\newblock In {\em Proceedings of the 19th Australasian Document Computing Symposium}, ADCS '14, page 58–65, New York, NY, USA. Association for Computing Machinery.

\bibitem[Vilar et~al., 2023]{vilar-etal-2023-prompting}
Vilar, D., Freitag, M., Cherry, C., Luo, J., Ratnakar, V., and Foster, G. (2023).
\newblock Prompting {P}a{LM} for translation: Assessing strategies and performance.
\newblock In Rogers, A., Boyd-Graber, J., and Okazaki, N., editors, {\em Proceedings of the 61st Annual Meeting of the Association for Computational Linguistics (Volume 1: Long Papers)}, pages 15406--15427, Toronto, Canada. Association for Computational Linguistics.

\bibitem[Xu et~al., 2024]{xu2024paradigm}
Xu, H., Kim, Y.~J., Sharaf, A., and Awadalla, H.~H. (2024).
\newblock A paradigm shift in machine translation: Boosting translation performance of large language models.

\bibitem[Ye and Li, 2023]{10.1007/978-981-99-7894-6_7}
Ye, N. and Li, J. (2023).
\newblock A k-nearest neighbor approach for domain-specific translation quality estimation.
\newblock In Feng, Y. and Feng, C., editors, {\em Machine Translation}, pages 69--80, Singapore. Springer Nature Singapore.

\bibitem[Yuan et~al., 2022]{Yuan2022AnIM}
Yuan, B., Li, Y., Chen, K., Lu, H., Yang, M., and Cao, H. (2022).
\newblock An improved multi-task approach to pre-trained model based mt quality estimation.
\newblock In {\em CCMT}.

\bibitem[Zerva et~al., 2022]{zerva-etal-2022-findings}
Zerva, C., Blain, F., Rei, R., Lertvittayakumjorn, P., C.~De~Souza, J.~G., Eger, S., Kanojia, D., Alves, D., Or{\u{a}}san, C., Fomicheva, M., Martins, A. F.~T., and Specia, L. (2022).
\newblock Findings of the {WMT} 2022 shared task on quality estimation.
\newblock In {\em Proceedings of the Seventh Conference on Machine Translation (WMT)}, pages 69--99, Abu Dhabi, United Arab Emirates (Hybrid). Association for Computational Linguistics.

\bibitem[Zhu et~al., 2023]{zhu2023multilingual}
Zhu, W., Liu, H., Dong, Q., Xu, J., Huang, S., Kong, L., Chen, J., and Li, L. (2023).
\newblock Multilingual machine translation with large language models: Empirical results and analysis.

\end{thebibliography}
\end{small}

% \FloatBarrier

% \newpage
\clearpage % Ensures the bibliography is on a separate page
\onecolumn

\appendix
\section*{Appendices}

% Overview

% \section{QE stats}
% QE stats

\begin{table}[H]
\centering
\begin{tabular}{lcc}
\toprule
\textbf{Metric} & \textbf{Generic Model} & \textbf{Specific Model} \\ 
\midrule
\textbf{Training Time (hh:mm)} & 05:55 & 06:54 \\ 
\textbf{CO2 Emissions (kg)} & 1.41 & 1.46 \\ 
\textbf{Electricity Consumption (kWh)} & 3.63 & 3.76 \\ 
\bottomrule
\end{tabular}
\caption{\textbf{Training Time, CO2 Emissions, and Electricity Consumption for QE Models.}}
\label{app:QEstats}
\end{table}

% \section{ICE example}
%ICE example

\begin{table}[H]
\centering
\renewcommand{\arraystretch}{1.1}
\scalebox{0.90}{
\begin{tabular}{lll}\hline \hline
\addlinespace[2pt]
\multicolumn{3}{l}{\textbf{\underline{ICEs:}}}  \\
\multicolumn{3}{p{15cm}}{Die Sockets, die im except Array aufgelistet sind, werden auf Ausnahmen überwacht. = The sockets listed in the except array will be watched for exceptions. $</s>$ Geben Sie den Namen der Variablen ein, deren Wert überwacht werden soll. = Enter the name of the variable whose value is to be monitored. $</s>$ Nur erlaubt bei Sockets für lokale Displays und den globalen Socket. = Permitted only on sockets of local displays and the global socket. $</s>$ Legt fest, ob Scandaten-Information, die in den MPEG2-Videoströmen enthalten sind, aktualisiert werden sollen. = This controls whether to update the scan data information contained in the MPEG-2 video streams. $</s>$ Die Sockets, die im write Array aufgelistet sind, werden daraufhin überwacht, ob ein Schreibvorgang den Socket blockiert. = } \\ \addlinespace[3pt] \hline
\addlinespace[2pt]
\multicolumn{3}{l}{\textbf{\underline{Translation:}}} \\
\multicolumn{3}{l}{The sockets listed in the write array will be watched for whether a write operation blocks the socket.} \\ \addlinespace[3pt] \hline
\addlinespace[2pt]
\multicolumn{3}{l}{\textbf{\underline{Reference Label:}}} \\
\multicolumn{3}{l}{The sockets listed in the write array will be watched to see if a write will not block.}  \\ \addlinespace[3pt] \hline
\addlinespace[4pt]
\textbf{QE:} 67.59, \textbf{BLEU score (using reference label):} 52.89 & 
\multicolumn{2}{c}{} \\ \addlinespace[3pt] \hline \hline
\end{tabular}}
\caption{\textbf{An example of selected ICEs} for a source text, its corresponding translation, reference label, and QE estimation compared to the BLEU score computed based on the reference label.}
\label{example of mode1}
\end{table}

\begin{figure*}[]
\begin{algorithmic}[1]
\Function{Search}{...}
    \State $\text{temp} \gets [(``", 0.0, ``")]$
    \State $\text{prompt} \gets ``"$
    \State $\text{itr} \gets 0$
    \State $\text{best\_qe\_score} \gets 0.0$
    \State $\text{patience\_counter} \gets 0$
    \While{$\text{itr} < \text{iteration} \text{ and } \text{patience\_counter} < \text{early\_stop\_patience}$}
        \State $\text{available\_Prompts} \gets \Call{GenerateAvailablePrompts}{...}$ \Comment{Initial ICEs}
        \If{$\text{available\_prompts}$ is not empty}
            \State $\text{selected\_prompt\_index} \gets \text{itr} \bmod k$ \Comment{Phase 1: Selection}
            \State $\text{selected\_prompt} \gets \text{available\_prompts}[\text{selected\_prompt\_index}]$
            \State $\text{prompt} \gets \Call{ConstructFullPrompt}{...}(see~\ref{LLM})$
            \State $\text{input\_ids[0]} \gets \Call{EncodePrompt}{...}$ \Comment{Phase 2: Translation}
            \If{$\text{length}(input\_ids) > \text{LLM\_max\_length}$}
                \State \Return $\text{temp}$
            \EndIf
            \State $\text{output} \gets \Call{GenerateOutput}{...}$ 
            \State $\text{final\_output} \gets \Call{DecodeOutput}{...}$
            \State $\text{qe\_input} \gets \Call{PrepareQEInput}{\text{source}, \text{final\_output}}$ \Comment{Phase 3: Estimation}
            \State $\text{qe\_score} \gets \Call{EstimateQuality}{\text{qe\_input}, \text{model\_QE}}$
            
            \State $\text{temp}.\text{append}((\text{prompt}, \text{current\_qe\_score}, \text{final\_output}))$
            \If{$\text{current\_bleu\_score} \geq 100$}
                \State \Return $\text{temp}$
            \EndIf
            \State $\text{temp} \gets \Call{SortTemp}{...}$
            \If{$\text{qe\_score} \leq \text{best\_qe\_score}$}
                \State $\text{patience\_counter} \gets \text{patience\_counter} + 1$
            \Else
                \State $\text{patience\_counter} \gets 0$
            \EndIf
            \State $\text{best\_qe\_score} \gets \text{temp}[0][1]$
        \EndIf
        \State $\text{itr} \gets \text{itr} + 1$
    \EndWhile
    \State \Return $\text{temp}$
\EndFunction
\end{algorithmic}
    \captionof{algorithm}{\textbf{Pseudocode outlining the proposed Search Algorithm.} Each phase of the methodology is annotated alongside the relevant code. Function arguments are omitted for simplicity. The first element of the returning list (\textit{temp}) includes the selected prompt, its associated QE score, and the translated text.}
    \label{search_pseudo}
\end{figure*}

\end{multicols}
\end{document}